%% file: main.tex
\documentclass[letterpaper, 10 pt, conference]{ieeeconf}  

\IEEEoverridecommandlockouts                              

\input{preamble}

\newcommand{\citep}[1]{\cite{#1}}
\newcommand{\citet}[1]{\cite{#1}}

\newcommand{\nhardpar}[0]{3}

\newcommand{\hardpar}[0]{\parameters^m}
\newcommand{\softpar}[0]{\parameters^c}

\newcommand{\ourmethod}[0]{HPC-BBO}
\newcommand{\BBO}[0]{BBO}
\newcommand{\CBO}[0]{cBO}
\title{\LARGE \bf
Data-efficient Learning of Morphology and Controller for a Microrobot
}

\author{Thomas Liao$^{1}$, Grant Wang$^{1}$, Brian Yang$^{1}$, Rene Lee$^{1}$, Kristofer Pister$^{1}$, Sergey Levine$^{1}$, and Roberto Calandra$^{2}$
\thanks{*This work was supported by Berkeley DeepDrive}
\thanks{$^{1}$ University of California, Berkeley, CA, USA\newline
        {\tt\small \{thomas.liao, grant.wang5, brianhyang, renelee, ksjp\}@berkeley.edu, svlevine@eecs.berkeley.edu}}%
\thanks{$^{2}$ Facebook AI Research, Menlo Park, CA, USA\newline
        {\tt\small rcalandra@fb.com}}%
}

\begin{document}

\maketitle
\begin{abstract}
	\input{0_abstract}
\end{abstract}


\section{Introduction}

	\input{1_introduction}


\section{Related Work}
\label{sec:related}

	\input{2_related}


\section{Problem Formulation}
\label{sec:formulation}

	\input{3_formulation}


\section{Background}
\label{sec:background}

	\input{4_background}


\section{Hierarchical Process Constrained Batch Bayesian Optimization}
\label{sec:approach}

\input{5_approach}


\section{Experimental Results}
\label{sec:results}

	\input{6_results}


\section{Conclusions}
\label{sec:conclusion}

	\input{7_conclusion}




\clearpage

\bibliographystyle{IEEEtran}

\input{main.bbl}
%
%

\end{document}

%% file: preamble.tex
\usepackage[english]{babel}

\usepackage{graphicx}
\usepackage{animate}
\usepackage{epsfig} 				
\usepackage{epstopdf}

\usepackage{listings}
\usepackage{color}
\usepackage{nameref}
\usepackage{hyperref}
\usepackage[colorinlistoftodos]{todonotes}
\usepackage{amsmath}	 			
\usepackage{amssymb}  				

\usepackage{dsfont}			
\usepackage{mathtools}

\usepackage{epigraph}
\usepackage{lscape}
\usepackage[]{nomencl}				
\usepackage{algorithm}
\usepackage{algpseudocode}
\algtext*{EndWhile}
\algtext*{EndIf}
\algtext*{EndFor}

\usepackage{multicol}
\usepackage{multirow}
\usepackage{etoolbox}

\usepackage{caption}
\usepackage{subcaption}

\usepackage{wrapfig}
\usepackage{multimedia}

\usepackage{siunitx}
\usepackage{flushend}

\usepackage{floatflt}

\usepackage{url}

\usepackage[absolute,overlay]{textpos}

\usepackage{bibentry}

\usepackage{tikz}
\usepgflibrary{arrows}	

\usepackage{float}
\usepackage[utf8]{inputenc}
\usepackage[english]{babel}
\usepackage{graphics}
\usepackage{animate}

\usepackage{listings}

\usepackage[nostamp]{draftwatermark}

\usepackage{extarrows}		
\usepackage{xcolor}
\definecolor{ultramarine}{RGB}{0,32,96}

\usepackage{changes}
\definechangesauthor[name={Thomas Liao}, color={blue}]{TL}
\definechangesauthor[name={Roberto Calandra}, color={orange}]{RC}

\allowdisplaybreaks[1]

\newcommand{\playvideo}[1]{\href{run:#1}{\includegraphics[scale=0.12]{\RCPath fig/empty}}}

\usepackage{lipsum}
\usepackage{framed}

\input{commands.tex}

\AtEndDocument{\par\leavevmode}

%% file: commands.tex
\newcommand{\ie}[0]{i.e.}

\newcommand{\wrt}[0]{w.r.t.}

\newcommand{\email}[1]{\href{mailto:#1}{\nolinkurl{#1}}}

\newcommand{\link}[1]{\colora{\url{#1}}}

\renewcommand{\sec}[1]{Section~\ref{#1}}
\newcommand{\fig}[1]{Figure~\ref{#1}}

\newcommand{\tab}[1]{Table~\ref{#1}}
\newcommand{\alg}[1]{Algorithm~\ref{#1}}

\definecolor{matlab1}{rgb}{0,0,1}
\definecolor{matlab2}{rgb}{0,0.5,0}
\definecolor{matlab3}{rgb}{1,0,0}
\definecolor{matlab4}{rgb}{0,0.75,0.75}
\definecolor{matlab5}{rgb}{0.75,0,0.75}
\definecolor{matlab6}{rgb}{0.75,0.75,0}
\definecolor{matlab7}{rgb}{0.25,0.25,0.25}

\definecolor{darkgreen}{rgb}{0,0.5,0}		
\definecolor{purple}{rgb}{0.75,0,0.75}
\definecolor{pink}{rgb}{1,0.4,0.6}

\newcommand{\capitalize}[1]{\expandafter\MakeUppercase\expandafter{#1}}

\newcommand{\colora}[1]{{\usebeamercolor[fg]{framesubtitle}#1}}

\makeatletter
\newcommand*{\compress}{\@minipagetrue}
\makeatother

\renewcommand{\vec}[1]{\boldsymbol{#1}}				

\newcommand{\R}[0]{\mathds{R}}					

\newcommand{\dataset}[0]{\mathcal{D}}

\newcommand{\maximize}[0]{\text{arg max}} 			

\newcommand{\parameter}[0]{\theta} 				
\newcommand{\parameters}[0]{\vec{\parameter}} 			
\newcommand{\context}[0]{\vec{c}}
\newcommand{\objfuncNo}[0]{f}					
\newcommand{\objfunc}[1]{\objfuncNo\left(#1\right)}		

\newcommand{\q}{\vec{q}}					
\ifdef{\dq}{\renewcommand{\dq}{\dot{\q}}}{\newcommand{\dq}{\dot{\q}}}

\usepackage{array}
\makeatletter
\newcommand{\thickhline}{%
    \noalign {\ifnum 0=`}\fi \hrule height 1pt
    \futurelet \reserved@a \@xhline
}
\newcolumntype{"}{@{\hskip\tabcolsep\vrule width 1pt\hskip\tabcolsep}}
\makeatother

%% file: 0_abstract.tex
Robot design is often a slow and difficult process requiring the iterative construction and testing of prototypes, with the goal of sequentially optimizing the design.
For most robots, this process is further complicated by the need, when validating the capabilities of the hardware to solve the desired task, to already have an appropriate controller, which is in turn designed and tuned for the specific hardware.
In this paper, we propose a novel approach, HPC-BBO, to efficiently and automatically design hardware configurations, and evaluate them by also automatically tuning the corresponding controller.
HPC-BBO is based on a hierarchical Bayesian optimization process which iteratively optimizes morphology configurations (based on the performance of the previous designs during the controller learning process) and subsequently learns the corresponding controllers (exploiting the knowledge collected from optimizing for previous morphologies).
Moreover, HPC-BBO can select a ``batch" of multiple morphology designs at once, thus parallelizing hardware validation and reducing the number of time-consuming production cycles.
We validate HPC-BBO on the design of the morphology and controller for a simulated 6-legged microrobot.
Experimental results show that HPC-BBO outperforms multiple competitive baselines, and yields a $360\%$ reduction in production cycles over standard Bayesian optimization, thus reducing the hypothetical manufacturing time of our microrobot from 21 to 4 months.

%% file: 1_introduction.tex
Designing intelligent robots to solve complex real-world tasks can be a daunting challenge.
The dominant paradigm is based on the flawed assumption that the morphology of a robot (\ie, the hardware) can to a large extent be designed independently of the underlying controllers.
In practice, this results in either designing general-purpose morphologies which can in theory solve a wide range of tasks, at the expense of being sub-optimal for any specific one, or in using an iterative process where each morphology design is followed by the design of an appropriate controller, with modifications (based on expert knowledge) to the previous morphology design to improve the chances of achieving a better controller at the next iteration.
Both of these approaches usually require a significant amount of expert knowledge which heavily influences the ultimate performance of the system.
Moreover, this process often requires a significant amount of design \emph{and} manufacturing time for each morphology and controller.

One alternative to this paradigm is to autonomously optimize both morphology and controller, based directly on the performance achieved on the specific application of interest.
This paradigm is conceptually similar to the evolutionary theory of Lamarckian inheritance~\citep{Lamarck1809,Koonin2009}, where the physical features of a species are directly guided through the generation by the ``usefulness'' of the features to the specific environment inhabited.
Although this evocative idea has already been proposed in past robotic
literature~\citep{Lipson2000,Jelisavcic2017}, the proposed approaches are difficult to apply to real-world applications due to the need for accurate analytical models, or a high number of evaluations to find good morphology/controller configurations. 

\begin{figure}[t]
  \centering
  \includegraphics[width=0.94\linewidth]{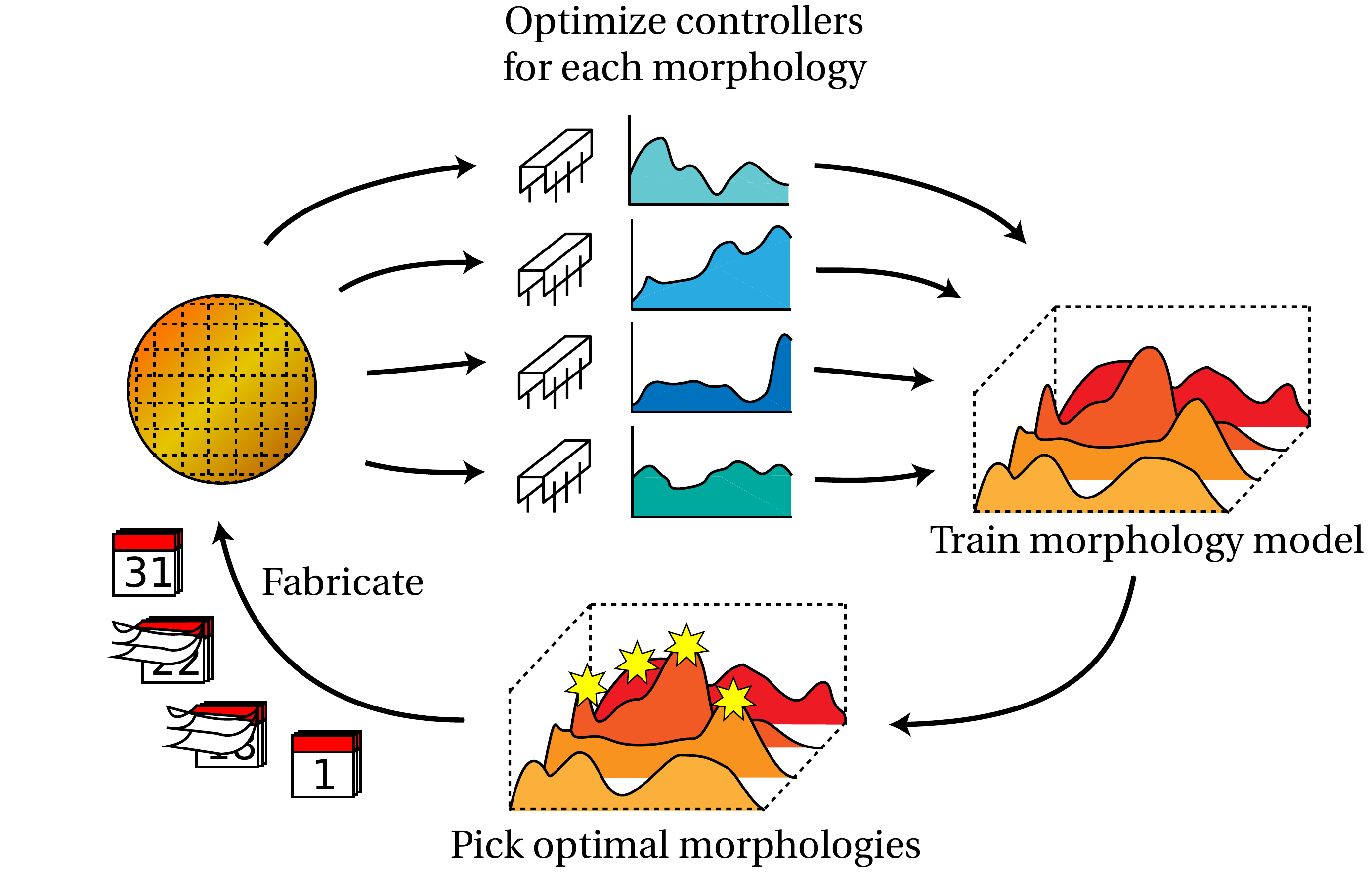}
  \caption{Our approach for learning morphology and controller is motivated by a setting where the manufacturing of a batch of microrobots is costly (both in terms of time and resources), but multiple robots can be produced in parallel on the same silicon waffer. Each robot can independently learn a controller and, at the end of the learning, a new batch of morphologies is selected to be manufactured next, based on the performance achieved by the current prototypes.}
  \label{fig:teaser}
\end{figure}
The motivating application of this paper is the design of a hexapod microrobot and its corresponding controller.
Due to the lack of sufficiently accurate models at the micro-scale, previous efforts~\cite{Contreras2017} have proven dependent on expert knowledge.
This robot is printed on silicon wafer, involving an expensive process with long production times (wafer delivery approximately 4-6 weeks after ordering).
With the current manufacturing process, each wafer has sufficient space to contain up to 5 printed robots.
After printing and assembly, it is still necessary to design a controller for the micro-robot, another design process largely guided by human experience.
Moreover, it is important to design this controller within a small number of experiments before the robot wears out.

As a result of these real-world manufacturing and design constraints, we focus in this paper on the need for an efficient optimization process, able to find suitable designs within an extremely limited number of iterations.
Hence, we investigate the following question: 
How can we design an algorithm for morphology/controller adaptation that is sufficiently data-efficient to be applicable to real-world robots?

Our contribution is two-fold: 
First, we propose a new optimization algorithm for co-optimization of robot morphology and controller in a data-efficient manner; Second, we validate this approach in simulation and show that it outperforms other state-of-the-art learning methods.
Our method exploits the hierarchical relationship between morphology and controller to produce optimal robots despite limited wall time. 
A sketch of this process is shown in \fig{fig:teaser}.
To demonstrate our method, we optimize the design  and controller of a recently developed hexapod  microrobot~\citep{Contreras2017}.
We use a simulation of this microrobot to validate our approach, but our method is not dependent on such a simulation existing. 
Indeed, the data-efficient nature of our approach allows application to the real-world microrobot.
Our approach is -- without loss of generality -- applicable to other robotic applications that benefit from joint morphology/controller optimization.
This in an exciting frontier towards enabling real-world robots that can quickly adapt both morphology and controller to perform specific tasks with high performance.    

%% file: 2_related.tex
\subsection{Bayesian Optimization}
	Bayesian optimization has been used to optimize both the design of mechanical systems~\citep{Martinez-Cantin2018}, as well as control policies~\citep{Lizotte2007,Tesch2011,Calandra2015a}, including those for microrobots~\citep{Yang2018}.
	Our work differs from these previous works because we jointly optimize hardware design \textit{and} control policies.
	While control policies are relatively cheap to optimize, new hardware designs incur substantial costs in both money and time.
	We propose a hierarchical batch contextual Bayesian optimization approach which identifies multiple promising hardware candidates at once (\ie{}, to be printed on the same silicon wafer) and leverages past evaluations when optimizing the control policy.
	Our approach yields significant performance improvements compared to standard Bayesian optimization, as demonstrated in \sec{sec:results}.

\subsection{Morphology Optimization}
	Our work is related to efforts to optimize controls for robots with non-fixed or reconfigurable morphologies~\citep{Yim2007, Bongard2013, Cully2015, Bongard2006, Romanishin2013}, where the controller must handle changes in the robot's physical configuration, either to cope with unanticipated damage~\citep{Cully2015,Bongard2006} or prepare for different tasks~\citep{Romanishin2013}.
	Notably, \citep{Nygaard2018} jointly evolve both morphological and gait parameters on a physical robot able to change its leg lengths dynamically.
	Application of this method is restricted to robots able to rapidly, repeatedly, and precisely alter their own morphology.
	This kind of online body modification is impossible for our microrobot because each change requires fabrication and assembly of a new robot. 

	Other works allow for changes in morphology between iterations, but not dynamic adjustments (\ie, the robot cannot adjust its morphology during simulation or the real-world).
	\citep{Geijtenbeek2013} include muscle routing in their optimization of gaits of bipedal creatures, allowing variation of muscle attachment points within bounded regions.
	Although they produce highly natural-looking gaits, it is unclear how muscle routing translates to a robot without an analogous concept of muscles. Furthermore, such aesthetic virtues are not relevant to our robot. 
	The detailed modeling of muscles and ligaments also entails a massive number of parameters to optimize -- up to thirty parameters for muscle physiology alone -- and a resultant increase in optimization time.
	In contrast, we only have \nhardpar{} hardware parameters to optimize. 
	
\subsection{Evolutionary Robotics}
	Substantial work in evolutionary robotics has focused on jointly evolving morphology and controls of virtual lifeforms~\citep{Bongard2003,Lipson2000,Funes1998,Sims1994}.
    However, evolving body plans from scratch rather than tuning existing designs increases optimization time substantially.
    Such works also often blur the lines between morphology and control parameters \citep{Cheney2014}, which is disadvantageous when morphological changes are orders of magnitude more costly to evaluate. 
    Our method produces highly performing robots with far fewer morphological changes when compared to evolutionary techniques. 

\subsection{Embodied Machines}
	Both \citep{Geijtenbeek2013} and \citep{Ha2017} optimize morphology and controls simultaneously but do not exploit the hierarchical relationship between morphology and control.
	Such optimization often leads to convergence of morphology before convergence of control \citep{Joachimczak2016,Lipson2016} and fails to adequately explore the space of morphology parameters.  
	Anecdotally, we observed this in experiments using regular Bayesian optimization, which frequently converged to a poorly performing morphology where only the front four legs touched the ground.

	One proposed explanation by \citep{Lipson2016} suggests that morphology mediates the role of the controller by functioning like an interface to the real world.
	Simultaneously changing morphology and control parameters is therefore counterproductive because each controller is specialized for some particular morphology.
	Early convergence of morphology is a consequence of heavy penalization of controller changes, since updating body plans negatively affects the performance of a controller optimized for a different body plan.
	In this vein, \citep{Luck2017} conduct policy search for several hardware schemes to learn optimal control-hardware combinations, running reinforcement learning for each hardware design they explore. 
	This work is most similar to ours because it jointly optimizes morphology and controls, but does not do so simultaneously and thereby avoids the early convergence problem.
	However, all the morphologies were bio-inspired and designed in advance by hand, whereas we include a large design space of morphologies in our optimization.
    Engineering morphologies by hand for our robot is notably harder because of a novel leg design that precludes straightforward transfer of other work \citep{Hoover2008,Wampler2009} in legged locomotion.

%% file: 3_formulation.tex
We formulate learning the morphology and controller of our microrobot as the optimization 
\begin{align}
	\parameters^* = \maximize_{\parameters}\, \objfunc{\parameters}\,,
	\label{eq:optimization}
\end{align}
of the parameters~$\parameters = [\hardpar,\softpar]$, where $\hardpar\in\R^n$ denotes the parameters of the morphology and $\softpar\in\R^m$ the parameters of the controllers, \wrt{} the desired objective function~$\objfuncNo$.

Although this problem can be solved as a single joint optimization task, changing the morphology at each optimization step is extremely costly for our microrobot as each fabrication takes up to a month.
Hence, it is crucial to minimize the amount of morphology evaluations.
On the other hand, once a morphology is available we can perform hundreds of controller evaluations at little cost.
This difference in evaluation cost already suggests that the formulation as a single joint optimization might not be desirable.

An alternative formulation is as a hierarchical optimization task with two independent levels of optimization, morphology optimization on top and controller below, where we alternate between selecting a morphology and optimizing the corresponding controller.
This formulation offers a natural way of decoupling the number of evaluations performed on the controller from the evaluations of the morphology.
A further improvement on this formulation is to consider the batch nature of the morphology evaluations for our application, selecting multiple morphologies to be manufactured (and later evaluated) at once.
One drawback of this hierarchical formulation is that decoupling the two levels of optimization prevents information sharing between them.
In practice, this means each controller optimization process cannot make use of information provided by previous controller optimizations, and thus needs to start from scratch.

Our approach, presented in \sec{sec:approach}, extends the hierarchical batch formulation and allows to make full use of data collected from previous controller optimizations, thus further improving the data-efficiency.

%% file: 4_background.tex
\subsection{Central Pattern Generators}
	Central pattern generators (CPGs) are neural circuits commonly found in vertebrates that do not need sensory input to produce periodic outputs~\citep{Yu2014}.
	They have been used widely in the design of gaits for robotic locomotion \citep{Ijspeert2008,Righetti2006,Crespi2008}. 
	We chose to use CPGs for our controller.
    For reasons why CPGs are a good choice of controller for microrobots, we refer readers to \citep{Yang2018}.
	The dynamics of CPGs are modeled as a network of coupled non-linear oscillators.
	For an in-depth explanation of how the oscillators in CPG networks work, we refer readers to~\citet{Crespi2008}.
    
	A major benefit of using CPGs for our controller is the low number of parameters~$\softpar$ to optimize.
	Usually, the parameters optimized are $\softpar = \left[\omega, R, X_{l}, X_{r}\right]$ where $\omega$ is the desired frequency of the oscillators, $R$ is the phase difference between each vertical-horizontal oscillator pairs, and $X_{l}$ and $X_{r}$ are the amplitudes of the left and right side oscillators and allow for directional control of the microrobot.

\subsection{Bayesian Optimization}
	One method often used to automate the parameter tuning process is Bayesian optimization~(BO).
	BO is a zero-order black-box optimizer often used for global optimization of expensive functions~\citep{Kushner1964, Jones2001}.
	At every iteration of the optimization, BO learns a model $\tilde{\objfuncNo}: \parameters \rightarrow \objfunc{\parameters}$ from the dataset of previously evaluated parameters and their returned objective values~$\dataset=\{\parameters, \objfunc{\parameters}\}$.
	The learned model is then used to execute a virtual optimization by using an acquisition function which controls the trade-off between exploitation and exploration. 
	The returned parameters $\parameters^*$ from this optimized model are then evaluated on the real system to obtain an objective value $\objfunc{\parameters^*}$.
	Finally, the parameters evaluated $\parameters^*$ and the corresponding objective value obtained from the real system $\objfunc{\parameters^*}$ are added to the dataset, and a new iteration of the optimization begins. 
	The choice of model is important for BO to learn the underlying objective.
	One commonly used model, and the one which we use in this paper, is the Gaussian process~(GP) model~\citep{Rasmussen2004}. 
	For a more in-depth background on BO, we direct readers to \citep{Jones2001, Kushner1964, Shahriari2016}.
    
    An extension of standard BO we use as a subcomponent is \emph{contextual Bayesian optimization}~(cBO)~\citep{Metzen2015}.
    cBO extends the standard BO framework by augmenting the optimization problem with an additional context parameter~$\context$, and learns a joint policy $\tilde{\objfuncNo}: \{\parameters,\context\} \rightarrow \objfunc{\parameters}$, where $\context$ is fixed during optimization (i.e., it is observable, but not controllable). 
    In our approach, we use cBO to optimize the controller, as described in \sec{sec:approach}.
    By encoding the morphologies as contexts, cBO takes advantage of the similarities between different morphologies and generalizes to good polices for unseen designs faster.

	Within our approach, we also use another variant of BO called \emph{batch Bayesian optimization}~(BBO)~\citep{Vellanki2017} to optimize the morphology. 
	In contrast to a fully sequential algorithm, which alternates between choosing individual points and evaluating them on the true reward function, BBO queries the acquisition function for multiple points, then evaluates them in parallel before selecting another. 
	The first set of parameters of each batch is selected as in a sequential policy and the next set chosen with an acquisition function. 
	However, rather than immediately evaluate the returned parameters on the real system, BBO defers evaluation of the reward function on this point until the entire batch is selected, temporarily substituting for its reward a prediction $\mathcal{H}$ made by the GP (also called \textit{hallucinated observations} \citep{Desautels2014} or \textit{fantasies} \citep{Shahriari2016}).
    The GP model is updated with the data point $\dataset=\{\parameters, \mathcal{H}\}$ of returned parameters and respective hallucinated observation and is then used to select the next point.
    Once the entire batch is selected, all points are evaluated and the hallucinated observations are replaced by real ones.
    Although hallucinated rewards are less informative, batching saves time since points in the batch can be evaluated in parallel.
    The particular implementation of BBO we use as a subcomponent in our approach is PC-BBO~\citep{Vellanki2017}.

%% file: 5_approach.tex
Our algorithm is called \textit{hierarchical process constrained batch Bayesian optimization}~(HPC-BBO). 
In this context, ``process-constrained'' refers not to classical constrained optimization, but rather to a physical limitation that restricts how frequently a particular parameter can be changed. 
In our case, it would be relatively straightforward to change control parameters (unconstrained) on a physical microrobot, compared to fabricating a new hexapod to test a different morphology (constrained).

\begin{algorithm}[t]
\caption{Hierarchical Process Constrained Batch Bayesian Optimization (HPC-BBO) -- contextual case}
\label{alg:pco algorithm}
\begin{algorithmic}[1]
    \State $\hardpar_{1:K}\leftarrow[\hardpar_{1}, ..., \hardpar_{K}]$ \Comment{Randomly initialize  batch}
    \While{$b < NumBatches$}
      \For{$k = 1, .., K$}
      \Comment{Learn controllers}
	\While{$i < NumIters$}
	    \State{$\parameters^{c*} \leftarrow \maximize_{\parameters^{c}}\, \text{GP-UCB}(\parameters^c, \parameters^{m}_{k} | \mathcal{G}_c)$}
	    \State $\objfunc{\parameters^{c*}, \hardpar_{k}} \leftarrow \parameters^{c*}$ \Comment{evaluate on real system}
	    \State $\mathcal{G}_{c} \leftarrow \{\parameters^{c*}, \objfunc{\parameters^{c*}, \parameters^{m}_{k}}\}$
	\EndWhile
	\State $\mathcal{R^*}_k \leftarrow \objfunc{\parameters^{c*}, \parameters^{m}_{k}}$ \Comment{Keep best reward}
	\State $\mathcal{G}_{m} \leftarrow \{\parameters^{m}_{k}, \mathcal{R^*}_k\}$
      \EndFor  
      
      \State $\mathcal{G'}_m \leftarrow
      \mathcal{G}_m$ \Comment{Temporary model with hallucinations}
      \For{$k = 1, .., K$} \Comment{Generate new morphologies}
       \State{$\parameters^{m*} \leftarrow \maximize_{\hardpar}\, \text{GP-UCB}( \hardpar| \mathcal{G'}_m)$}
       \State{$\parameters^m_k \leftarrow  \parameters^{m*}$} \Comment{Add to batch}
       \State{$\mathcal{H^*} \leftarrow h(\parameters^{m}_{k} | \mathcal{G}_{m})$} \Comment{Hallucinate a reward}
       \State $\mathcal{G'}_{m} \leftarrow \{\parameters^{m*}_{k}, \mathcal{H^*}\}$ 
      \EndFor
    \EndWhile
\end{algorithmic}
\end{algorithm}

We now detail \alg{alg:pco algorithm}.
Given a set of morphology parameters~$\hardpar$ and controller parameters~$\parameters^{c}$, we want to jointly optimize them.  
We use \BBO{} to select the morphology parameters~$\hardpar$ and evaluate them by having a nested  optimization procedure for the controller parameters~$\softpar$.
At the end of the controller optimization, we return to the morphology optimizer the best reward obtained for that specific morphology.
We initialize a batch of size $K$ by picking morphology parameters with random search.
In the contextual case, for each of step $1, \ldots, K$ of the current batch, we set the morphology parameters as a context, and perform contextual Bayesian optimization over the controller parameters using a GP model $\mathcal{G}_{c}$ that learns a policy $\tilde{\objfuncNo}: \{\parameters^{c},\context\} \rightarrow \objfunc{\parameters^{c}}$, where $\context=\parameters^{m}_{k}$.
The noncontextual variant of our algorithm uses standard BO instead of \CBO{} to optimize the controller, which necessitates training a new GP for each morphology in the batch for a total of $K$ controller GPs per batch.
Both the contextual GP and noncontextual GPs must first be initialized with randomly chosen controller parameters.
After the first batch is evaluated, another GP model $\mathcal{G}_{m}$ is used to learn a policy $\tilde{\objfuncNo}: \parameters^{m} \rightarrow \objfunc{\parameters^{c}}$ and updated with the batch and corresponding best rewards for each batch element evaluated from the controller optimization.
We query this updated model $\mathcal{G}_{m}$ using an acquisition function (GP-UCB in our case) for another batch of morphology parameters, this time with updated knowledge of what parameters generated the best rewards from the software optimization, and evaluate this batch as before.
Regardless of whether noncontextual~\ourmethod{} or contextual~\ourmethod{} is employed, the use of \BBO{} to optimize morphology allows all controller optimizations for a batch to be done in parallel.
By evaluating multiple morphologies in one batch, we drastically reduce wall time for optimization since we can evaluate $K$ morphologies in one production cycle, whereas regular BO evaluates one. 
In essence, contextual \ourmethod~leverages the data efficiency of \BBO{} to optimize the expensive constrained parameters while also taking advantage of the information learned across different contexts with \CBO{} to optimize the unconstrained parameters.
Using our approach, we can co-optimize robot control and morphology in a much more data-efficient manner, allowing us to evaluate more microrobots per fabrication cycle.

%% file: 6_results.tex
\subsection{Experimental Setting}
\begin{figure}
  \centering
  \includegraphics[height=2.9cm]{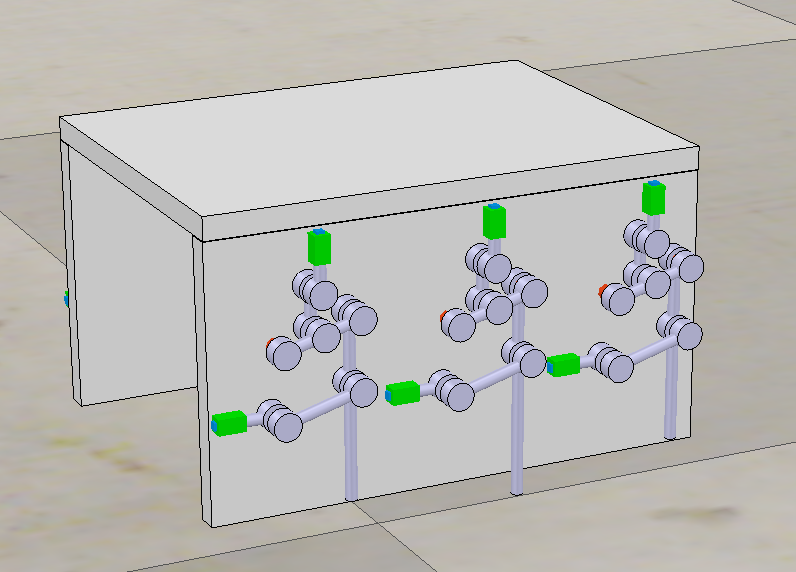}
  \hfill
  \includegraphics[height=2.9cm]{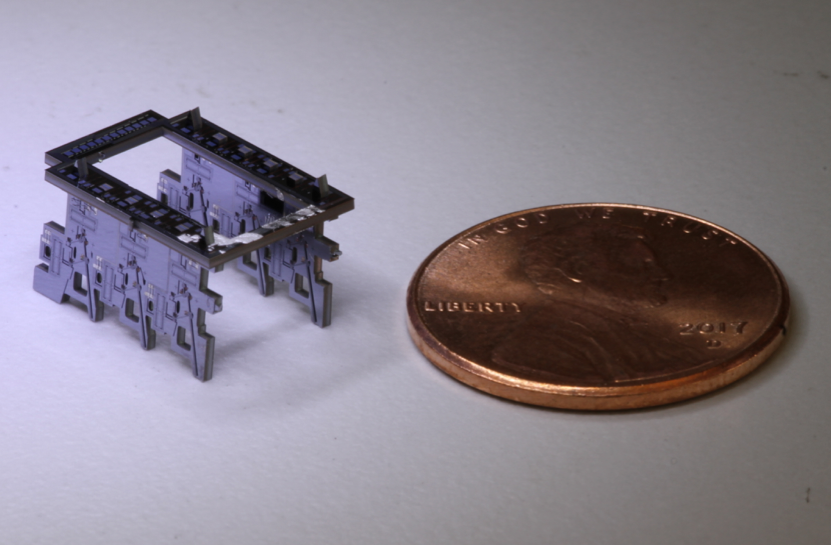}
  \caption{The simulated (\textit{left}) and real microrobot walker (\textit{right}) considered in our work.}
  \label{fig:microrobot}
\end{figure}
\begin{figure*}[t]
  \centering
  \includegraphics[width=\linewidth]{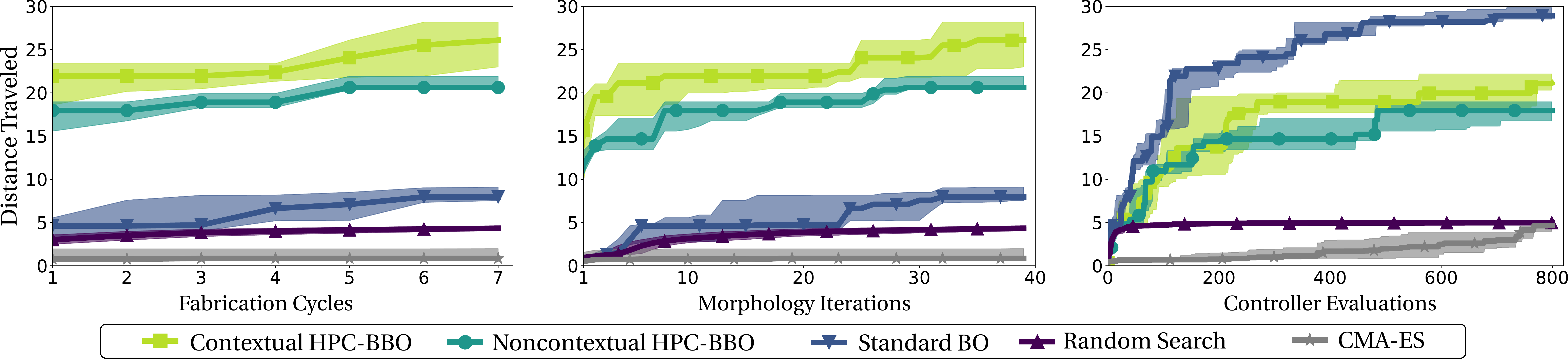}
  \caption{Learning curves (median and 65th percentile of the best performance observed so far) for multiple algorithms w.r.t. (\textit{left}) number of manufacturing batches, (\textit{center}) number of morphologies evaluated, and (\textit{right}) number of experiments performed.
  Each fabrication cycle corresponds to a wafer of 5 morphologies.
      Although standard BO performs better than our approach w.r.t. the pure number of evaluations, when considering the number of fabrication cycles and number of morphologies evaluated, our approach significantly outperforms all other baselines, with a $267\%$ improvement over standard BO for the non-contextual version and a $360\%$ improvement over standard BO for the contextual version.
  In the real fabrication process of the microrobot, where the fabrication cycle of a wafer takes about a month, this would correspond to 4 months for our approach and 21 months for standard BO to reach the same performance.}
    \label{fig:optimization}
\end{figure*}
	In our experiments, we used the robotic simulator V-REP to model a hexapod microrobot similar to the one described in~\citep{Yang2018}. 
	Each of the six legs is driven by 2 motors: one motor actuates the leg vertically, and the other motor moves the leg back and forth -- resulting in the legs having a circular-like sweep. 
	The simulated microrobot is scaled 100 times larger than its physical analogue since V-REP is unable to handle dynamics at the micrometer scale. 
	It is important to notice that our algorithm does not require a simulator to work, and that the simulator is here used only to evaluate the performance of the robots, similar to what would happen in the real world.
    
    For the controller, we optimize six CPG parameters, which correspond to the frequency, amplitude, and offsets of the vertical and horizontal motors.
    In addition, we separately consider parameters that control the amplitude of leg swings on the left and right sets of legs respectively.
    Although most of the parameters are related to our controller of choice, the CPG, our method is agnostic to the type of controller.
    The three morphology parameters control the lengths of pairs of legs (front, middle, and rear) and are encoded as ratios relative to a normalized leg length.  
    We use a batch size $K=5$ and run the controller optimizer for 50 iterations for each morphology evaluation.
    Videos and code for reproducing the experiments are available at \url{https://sites.google.com/view/learning-robot-morphology}

\subsection{Comparison to Other Methods}

          We now compare the two variants of our approach (contextual \ourmethod{} and non-contextual \ourmethod) against three baselines: random search, covariance matrix adaptation evolutionary strategy (CMA-ES), and standard BO.
          Random search~\citep{Brooks1958} samples the parameter space as a uniform distribution, establishing a baseline for our optimization task.
          Covariance matrix adaptation evolutionary strategy is a gradient-free algorithm to optimize non-convex functions~\citep{Hansen2016}. 
          Standard BO optimizes all parameters at once and, unlike our hierarchical approach, does not batch hardware evaluations.

          \fig{fig:optimization} shows the learning curves of the various methods w.r.t. different optimization desiderata: number of fabrication cycles, morphology iterations, and controller evaluations.
          The number of fabrication cycles is number of times a new silicon wafer has to be fabricated and is the most important statistic for comparison because it directly correlates to wall time.
          By generating a batch of multiple morphologies, \ourmethod~examines many different sets of hardware parameters at each fabrication cycle, whereas CMA-ES and standard BO only evaluate one new morphology every cycle. 
          The fabrication process of each batch of morphologies takes 4-6 weeks in the real world. 
          As a result, standard BO would need 21 months to reach the same performance that our approach would reach in 4 months (assuming that the convergence rate in simulation would translate to real world).
          \ourmethod{} also outperforms standard BO and CMA-ES w.r.t. the number of morphology iterations.
          This metric is relevant because it shows that even if the batch size $K=1$, \ourmethod{} would still outperform standard BO (since the larger batch sizes decrease performance, as explained in~\sec{sec:background}).

          \ourmethod{} is significantly outperformed by standard BO when considering the number of controller evaluations, because standard BO is able to see far more morphologies as it is able to change both controller and morphology parameters at the same time.
          However, since even just 200 controller evaluations would take a decade and a half, comparing \ourmethod{} to standard BO on this basis is misleading.

\begin{figure*}[t]
  \centering
  \includegraphics[width=0.98\linewidth]{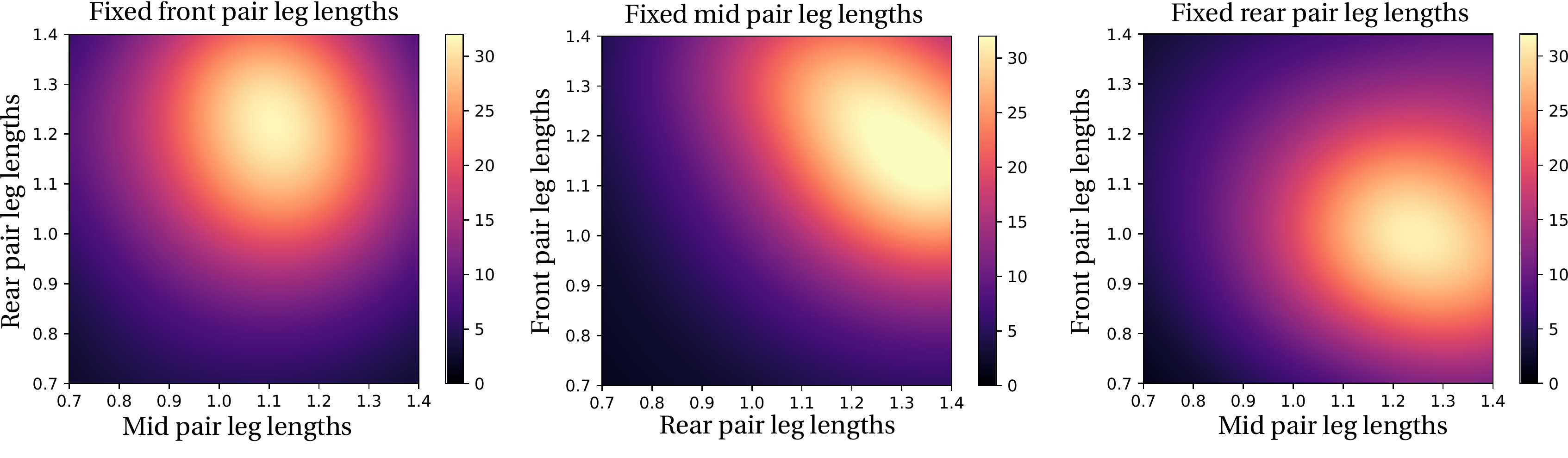}
  \caption{Visualization of the learned GP model for the hardware parameter space from contextual \ourmethod{}.
      The colormap indicates the distance walked, where 30 is far and 0 is stationary.
    We see from the left map that fixing the length of the front pair of legs to be equal to that of a specific high performing morphology, it is best paired with medium length middle legs and long rear legs.
      When we fix the middle leg as seen in the center map, it is best paired with long rear legs and long front legs.
      The right map indicates that when we fix the rear leg, it is best paired with long middle legs and short front legs.
  }
  \label{fig:model}
\end{figure*}
\begin{table}[t]
      \begin{center}
      \renewcommand{\arraystretch}{1.3}
      \centering
      \begin{tabular}{|c||c|c|c|c|}
      \hline
      \bfseries Morphology & \bfseries 1 & \bfseries 2 & \bfseries 3 & \bfseries 4\\
      \hline \hline
      Controller 1 & -- & -18.73\% & -7.35\% & -46.29\%\\
      \hline
      Controller 2 & -74.79\% & -- & -78.38\% & -29.34\%\\
      \hline
      Controller 3 & -75.52\% & -13.26\% & -- & -56.19\%\\
      \hline
      Controller 4 & -88.43\% & -65.83\% & -89.04\% & --\\
      \hline
      \end{tabular}
      \end{center}
    \caption{
          Changes in performance for a morphology/controller pair as a percentage of the reward, when changing the controller for a given morphology.
          The significant decreases in performance show that our hierarchical approach optimizes the best controller for each robot morphology instead of attempting to find a controller which works well for all morphologies. }
    \label{table:mc_table}
\end{table}

          The importance of our hierarchical technique is highlighted by the result presented in~\tab{table:mc_table}, which shows that the gaits we learn are specific to different morphologies. 
          We take the controllers and morphologies of the best four performing robots across all experiments and show that recombining them produces suboptimal pairings. 
          Highly performing morphologies can perform up to 90\% worse when paired with a controller optimized for another morphology, even when the other morphology-controller pair also performs well.
          Importantly, this relationship is not symmetric: even though the controller optimized for morphology~1 only decreases the performance of morphology 3 by 8\%, the controller for morphology~3 decreases the performance of morphology 1 by 75\%.

\begin{wrapfigure}{r}{0.5\linewidth}
  \centering
  \includegraphics[width=0.98\linewidth]{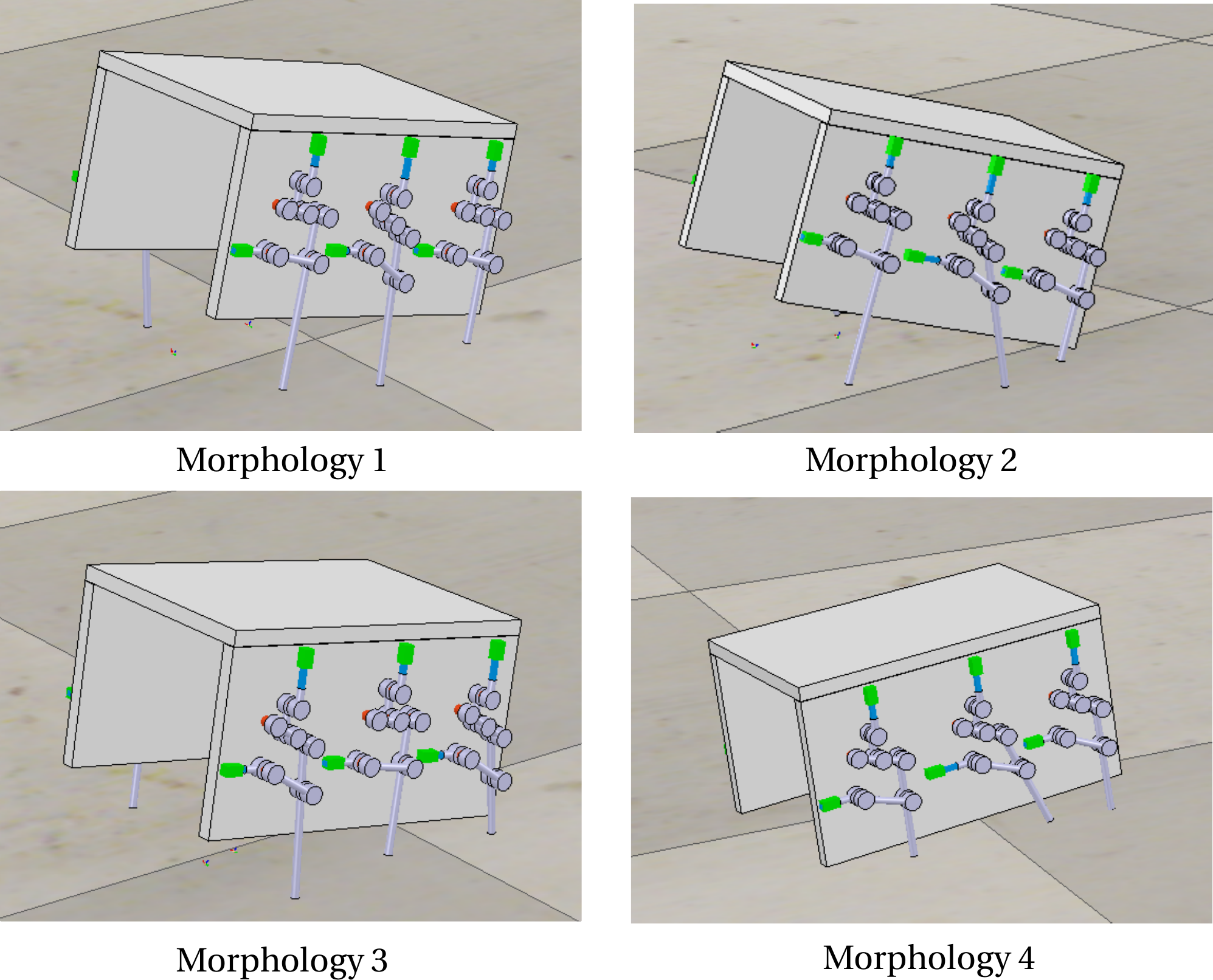}
  \caption{The high-performing morphologies from \tab{table:mc_table}.
      While morphology 4 has all short legs, the other morphologies have varied lengths and a tilted stance.
This may explain why other controllers perform so poorly when paired with morphology 4. 
  }
  \label{fig:bestmorph}
\end{wrapfigure}
 Even the controller that least decreases the performance of morphologies it was not optimized for, still decreases the performance of the morphology which has the least change in performance with different controllers.
          This means that even if standard BO or CMA-ES were to find a controller that performs well across most morphologies, they are not guaranteed to find the optimal controller/morphology pair.
          
          From \fig{fig:bestmorph}, we can see how the tilted stances of the morphologies obtained with \ourmethod{} resemble the inclined poses adopted by 6-legged insects in nature~\citep{Theunissen2015}.

\subsection{Contextual vs Non-contextual \ourmethod}
    We evaluate both the contextual and noncontextual variants of \ourmethod. 
    Contextual \ourmethod{} consistently outperforms noncontextual \ourmethod.
     As seen in \fig{fig:optimization}, the performance gap in distance traveled is largest at the beginning and towards the end w.r.t. the number of fabrication cycles and morphology iterations.
    Contextual \ourmethod{} outperforms non-contextual \ourmethod{} at the beginning of optimization because it is able to efficiently use data accumulated in previous morphology contexts to produce competitive gaits in unseen morphology contexts.
    This gap becomes less evident when the number of controller evaluations increases, as seen in the rightmost graph of \fig{fig:optimization}.
    As non-contextual \ourmethod{} trains on more morphologies, it picks better ones, which lessens the importance of generalizing across different contexts.
    This is because a higher number of controller evaluation iterations devoted to optimizing the software parameters allows the non-contextual optimizer to catch up to the contextual optimizer.
    With more iterations to optimize software parameters, the information of the contextual information from previous iterations is washed out, reducing the advantage the contextual optimizer has with fewer iterations. 
    The practical implication is that a shorter period of time to optimize software favors the contextual approach, whereas a longer one reduces its advantage.

%% file: 7_conclusion.tex
In this paper, we studied how to automatically optimize the design of robot morphologies and controllers in a data-efficient manner.
To achieve this goal, we introduced a novel algorithm, hierarchical process constrained batch Bayesian optimization~(\ourmethod{}), and validated our approach in simulation.
Results on a simulated hexapod microrobot show that \ourmethod{} significantly outperforms all other baselines and other state-of-the-art learning methods, with a performance improvement of $360\%$ over standard Bayesian optimization.
By exploiting the hierarchical relationship between morphology and controller, we demonstrate that \ourmethod{} can produce high-performing morphologies/controllers in a data-efficient manner.
Moreover, \ourmethod{} can exploit the simultaneous fabrication of multiple robot morphologies.
As a result, \ourmethod{} achieve the same performance of standard BO in a fifth of the time ($4$ months compared to $21$ months).

The proposed approach is a first step towards the grand goal of allowing robots that can not only quickly learn suitable controllers from experience, but also to adapt their hardware based on the needs dictated by their environment and goals.

An exciting future direction is to ``open the black-box'' by replacing Bayesian optimization with model-based reinforcement learning~\citep{Chua2018}, to allow for more complex controllers.
Additionally, we aim to apply the proposed approach to the design of real-world micro-robots.